\documentclass[letterpaper]{article} 
\usepackage{aaai25}  
\usepackage{times}  
\usepackage{helvet}  
\usepackage{courier}  
\usepackage[hyphens]{url}  
\usepackage{graphicx} 
\usepackage{natbib}  
\usepackage{caption} 
\usepackage{algorithm}
\usepackage{algorithmic}

\usepackage{booktabs}
\usepackage{multirow}
\usepackage{amsmath}

\usepackage{newfloat}
\usepackage{listings}
\DeclareCaptionStyle{ruled}{labelfont=normalfont,labelsep=colon,strut=off} 
\floatstyle{ruled}
\newfloat{listing}{tb}{lst}{}
\floatname{listing}{Listing}
\title{NomicLaw: Emergent Trust and Strategic Argumentation in LLMs During Collaborative Law-Making }
\author{
    Asutosh Hota\thanks{Corresponding author. Email: asutosh.jyu.hota@jyu.fi}, 
    Jussi P.P. Jokinen
}
\affiliations{
    University of Jyvaskyla, Faculty of Information Technology \\
    P.O. Box 35 (Agora), FI-40014 University of Jyvaskyla, Finland \\
    \{asutosh.jyu.hota, jussi.p.p.jokinen\}@jyu.fi

}

\usepackage{bibentry}

\begin{document}

\maketitle

\begin{abstract}
Recent advancements in large language models (LLMs) have extended their capabilities from basic text processing to complex reasoning tasks, including legal interpretation, argumentation, and strategic interaction. However, empirical understanding of LLM behavior in open-ended, multi-agent settings especially those involving deliberation over legal and ethical dilemmas remains limited.
We introduce \textbf{NomicLaw}, a structured multi-agent simulation where LLMs engage in collaborative law-making, responding to complex legal vignettes by proposing rules, justifying them, and voting on peer proposals. We quantitatively measure trust and reciprocity via voting patterns and qualitatively assess how agents use strategic language to justify proposals and influence outcomes.
Experiments involving homogeneous and heterogeneous LLM groups demonstrate how agents spontaneously form alliances, betray trust, and adapt their rhetoric to shape collective decisions. Our results highlight the latent social reasoning and persuasive capabilities of ten open-source LLMs and provide insights into the design of future AI systems capable of autonomous negotiation, coordination and drafting legislation in legal settings.
\end{abstract}

\begin{links}
  \link{Supplementary Materials (Code, Data \& Appendix)}{https://github.com/asutosh7hota/NomicLaw}
\end{links}

%

\section{Introduction}

\begin{figure*}[t]
    \centering
    \includegraphics[width=0.85\textwidth]{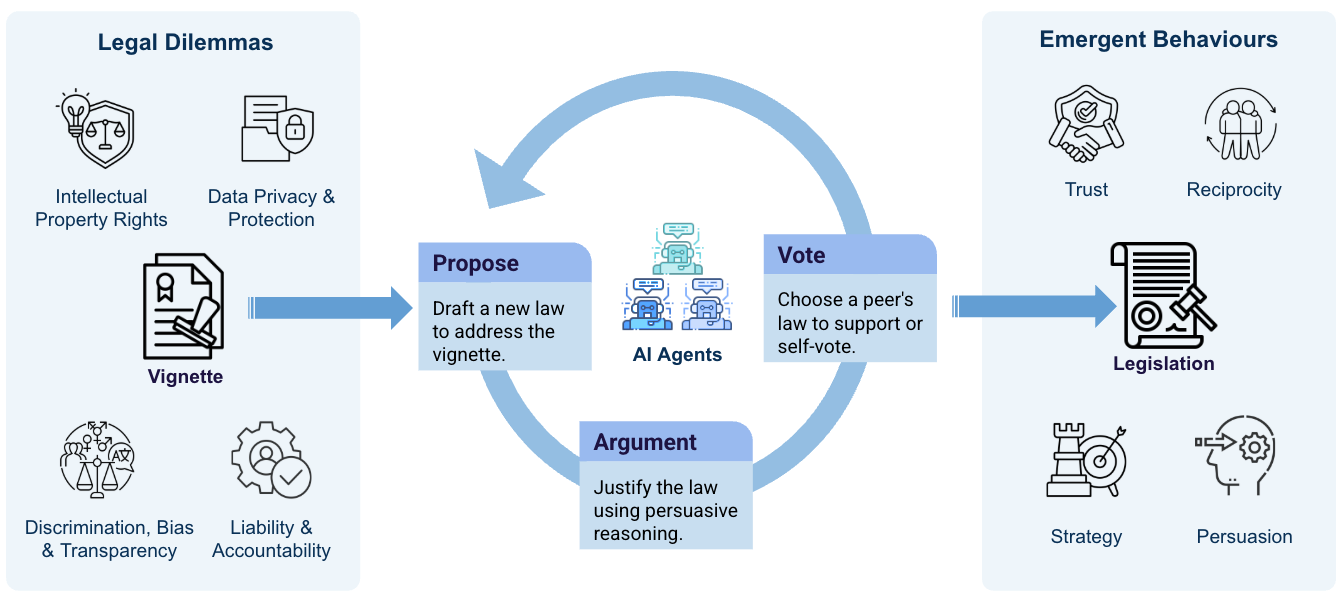}
    \caption{\textbf{NomicLaw framework:} This figure illustrates the core interaction loop in our Nomic-inspired legal simulation. Each game begins with a complex vignette that poses a legal dilemma. Agents respond by \textit{proposing} new laws to address the vignette, providing \textit{arguments} to justify their proposals, \textit{voting} on peer proposals, including their own and  \textit{legislating} with iterations. Over multiple rounds, agents accumulate scores based on whether their proposals are accepted, leading to strategic behavior. We observe emergent social dynamics such as persuasion, trust, reciprocity, and alliance formation across both homogeneous (same-model) and heterogeneous (multi-model) agent groups.}
    \label{fig:nomiclaw-framework}
\end{figure*}

Large language models (LLMs) now produce text that mimics human reasoning and deliberation, raising questions about their potential and risks in legal and legislative processes. While some studies suggest LLMs show emergent behaviors in negotiation and strategy \citep{brown2022human, piatti2024cooperate}, robust empirical evidence of their performance and limitations in complex, open-ended legal settings remains sparse ~\citep{castagna2024computational}. As AI systems become more visible in legislative support roles, it is crucial to rigorously assess both their capabilities and their fundamental shortcomings, particularly the risk of over-attributing genuine reasoning or judgment to statistical pattern matching \cite{west2023generative}.

To address this gap, we introduce \emph{NomicLaw}, an open-source multi-LLM framework for collaborative lawmaking (Figure~\ref{fig:nomiclaw-framework}). Drawing on ten state‐of‐the‐art models, NomicLaw embeds agents in a simplified propose–justify–vote loop over legally grounded vignettes, allowing them to draft rules, argue their merits and vote under a uniform point‐based incentive. This iterative sandbox yields emergent behaviors such as trust, reciprocity, and strategic persuasion and provides a framework for studying how model diversity shapes alliance formation, influence dynamics, and rhetorical efficacy in AI‐mediated lawmaking.

We test NomicLaw in two modes. In homogeneous sessions, five copies of the same model negotiate among themselves, revealing each models’ baseline tendencies leading to tighter alliances, frequent self-support and limited debate. In heterogeneous sessions, 10 distinct LLM agents interact, producing statistically significant shifts in behavior: models self-promote far less and form alliances that are roughly a quarter more variable, signaling richer, less insular deliberation. We also uncover clear performance gaps, for example, DeepSeek-R1’s proposals prevail most often, followed by Llama2, while others rarely succeed illustrating how agent diversity both broadens the scope of debate and exposes individual model strengths and weaknesses in collaborative rule-making. By open-sourcing NomicLaw, we provide a flexible platform for systematic exploration of multi-agent legislative tasks and for extending to human-in-the-loop legislative simulations, paving the way to understand hybrid human–AI collaboration in law making. Our contributions are:
\begin{enumerate}
  \item \textbf{A novel lawmaking sandbox.} An extendable propose–justify–vote protocol that elicits emergent alliances, trust dynamics, and strategic persuasion without fine‐tuning or role prescriptions.
  \item \textbf{A rich analytical framework.} Quantitative interaction metrics (e.g. self‐vote rate, coalition switch rate, reciprocity index etc.) paired with a hybrid LLM–human thematic analysis grounded in jurisprudence theory.
  \item \textbf{A reproducible toolkit:} Public release of code, prompt templates, simulation logs, and analysis scripts to catalyse further research on AI‐mediated governance, debate moderation, and policy co‐drafting. 
  \item \textbf{Insights into emergent behaviors.} Empirical characterization of strategic archetypes including win‐rate correlations and clustering analyses, revealing which LLMs are most persuasive or opportunistic.
\end{enumerate}

\section{Background}

Automated legal reasoning tasks have seen significant advances with the advent of LLMs. Early studies explored whether these models could replicate human-like legal inference in statutory contexts for example, GPT-4’s performance on syllogistic tasks \cite{spaic2024artificial} and the ability to generalize inference across legal domains \cite{marcos2024can}. Retrieval-augmented pipelines have been developed to better handle U.S. statutes and case law \cite{shu2024lawllm, yue2024lawllm}, and large-scale benchmarks now highlight both the strengths and persistent consistency gaps of LLMs in legal reasoning \cite{guha2023legalbench, jiang2023legal}. While prompt engineering can elicit displays of reasoning, persuasion, and negotiation from these models \cite{bassi2024decoding, kwon2024llms}, there are fundamental limitations: LLMs often reproduce patterns from their training data rather than engage in genuine, context-sensitive legal analysis. Unlike human judges, who synthesize facts with legal principles and may strategically frame arguments, LLMs' apparent legal reasoning frequently reflects statistical mimicry of past judicial data rather than true application of rules to novel legal cases \cite{doyle2025if}. This limits their reliability and interpretability in complex or unseen legal scenarios.

Scaling up LLM size does not consistently yield proportional improvements in legal reasoning. Evidence suggests diminishing returns on domain-specific benchmarks as model parameters increase \cite{diaz2024scaling}, and LLMs often mishandle advanced jurisprudential concepts, such as constitutional tests, without explicit prompts or guidance \cite{bignotti2024legal}. The opacity of LLM outputs further threatens due process in legal contexts. To address this, jurisprudential criteria for transparency have been proposed \cite{fresz2024should}, and recent hybrid models incorporate structured justification modules to improve interpretability \cite{ujwal2024reasoning, zhou2024analyzing}. While prompt engineering and stepwise prompting can enhance the appearance of legal reasoning, especially with domain-specific guides \cite{doyle2025if}, most multi-agent LLM simulations to date depend on prescribed roles or fine-tuning, limiting the spontaneous emergence of social behaviors and argumentation \cite{schneider2025generative}.

Parallel multi-agent AI research has examined negotiation, coalition building, and strategic interaction in domains such as resource allocation and trading \cite{piatti2024cooperate}. While social dynamics like trust, reciprocity, and alliance formation are well studied in psychology \cite{guo2025socialjax}, they are still largely unexplored in autonomous, LLM-driven legal settings. Computational argumentation has mostly addressed structured or static scenarios \cite{castagna2024computational}, often missing the fluid back-and-forth of claims and counterclaims that real legal collaboration requires. Although efforts to equip LLMs for structured argumentation such as retrieval-augmented synthesis \cite{gray2025generating}, hybrid symbolic-neural frameworks \cite{wang2024legalreasoner}, and moral heuristic embedding \cite{almeida2024exploring}—have advanced the field, these methods typically confine LLMs to fixed roles or isolated exchanges, failing to capture truly emergent or interactive behaviors.

Despite extensive research on reasoning, explainability, and argumentation, there is a notable lack of work investigating open-ended, multi-agent law-making by autonomous LLMs. To fill this gap, we present \emph{NomicLaw}, inspired by the self-amending game Nomic \citep{suber1982nomic}. Unlike prior studies that constrain agents to static tasks or pre-defined roles, NomicLaw provides a sandbox for LLM agents to collaboratively draft, negotiate, and iteratively revise legal rules in unconstrained, multi-agent legal deliberation.

\section{Methodology}

\subsection{NomicLaw Framework}
We developed NomicLaw, a flexible, multi-agent simulation framework in which LLM agents engage in a structured turn-based lawmaking game. Each game continues for five rounds per vignette. In each round, every agent independently proposes a new legal rule addressing the current vignette, justifies that proposal, and votes for exactly one proposal (self-voting is permitted) and briefly explains the rationale for the vote. Agents have no preset ideologies or private utilities beyond a simple point incentive: 10 points are awarded for a winning proposal and 5 points each for an undecided or tied vote. This mechanism guides agents to maximize acceptance of their own rules. All agents share full visibility of prior proposals, votes, justifications, and cumulative scores.

\subsection{Vignettes and Legal Domains}

We focus on four vignettes, each reflecting a distinct regulatory dilemma in AI governance. While these scenarios highlight important tensions in lawmaking, our approach does not capture the full procedural or political complexity of real-world legislative, regulatory, or judicial processes. Instead, the vignettes serve as testbeds for exploring how LLM agents deliberate over legal norms in a simplified, abstracted setting. These ensure both thematic breadth and direct relevance to ongoing regulatory debates. Table 1 (see Appendix) summarizes the scenarios and the legal domains they target. 

\subsection{Simulation Methodology}

We conducted experiments under two configurations: \textbf{homogeneous} and \textbf{heterogeneous}. In both, agents iteratively participated in five propose-and-vote rounds, leveraging a conversation buffer memory to retain full dialogue context. Each run was prompted by one of four legally charged vignettes.
In the homogeneous setting, we instantiated groups of five agents all using the same LLM for each of ten open-source models. Each model underwent one run per vignette, yielding 1000 observations in total.
In the heterogeneous setting, a single population of ten agents each powered by a distinct LLM was evaluated across six runs per vignette (24 runs total), producing 1200 agent-round observations. Agents scored ten points for a winning proposal and five points for ties or unanimous self-votes; all outcomes were logged as JSON entries detailing proposals, votes, winners, and scores. Agent memories were reset at the start of each run to prevent cross-run information leakage. After simulation, we exported logs by parsing filenames for each run, with the corresponding vignette id, round number, agent id, model, and voting patterns. The resulting CSV contained balanced entries (equal rounds per agent), which we verified programmatically before statistical analysis.

\subsection{Agent Configurations}

We compare two configurations to isolate the impact of model heterogeneity on emergent behavior.  In the \textit{homogeneous} condition, all agents in a single game instance run the same underlying model; in the \textit{heterogeneous} condition, each agent is backed by a different model.  We draw from a shared pool of ten open‐source LLMs (\texttt{phi4}, \texttt{phi4-reasoning}, \texttt{phi4-mini-reasoning}, \texttt{gemma3}, \texttt{gemma2}, \texttt{llama3}, \texttt{llama2}, \texttt{qwen3}, \texttt{granite3.3}, and \texttt{deepseek-r1}) were orchestrated through the Ollama API with identical settings (eg. temperature, standardized system prompt, retry logic).  By holding invocation parameters constant, we attribute differences in strategic adaptability and persuasive success solely to variations in model architecture, pre-training regimen, and alignment tuning.

\subsection{Exploratory Quantitative Metrics}

We extract the following metrics from our simulations: self-voting rate (SVR), average votes received (AVR), win rate (WR), vote volatility (VV), vote persistence (VP), reciprocity index (RI), coalition switch rate (CSR), bloc stability (BS), and edge density (ED). These metrics describe individual bias, group cohesion, decision volatility, and network connectedness. Formal definitions and empirical results for each metric appear in the Results section. Together, these exploratory metrics provide a quantitative foundation for comparing interaction patterns across homogeneous and heterogeneous model configurations.

\subsection{Qualitative Metrics}

We complement quantitative measures by computing agent influence and peer‐alignment (how proposals and votes sway subsequent positions), followed by a hybrid LLM–human thematic analysis: labelling 2,200 proposals and justifications with jurisprudential themes via two LLMs (10\% human‐validated, $\kappa \ge 0.7$), and assessing proposal–vote thematic consistency. This mixed approach reveals both the interaction dynamics and the normative frames driving consensus.

\subsection{Reproducibility and Extensibility}
All code, prompt templates, and anonymized log data are publicly released to ensure full reproducibility.  NomicLaw’s architecture allows researchers to seamlessly integrate new vignettes, adjust group sizes or round counts, introduce human participants or varied incentive structures, and plug in custom evaluation modules (e.g.\ expert ratings, fact‐checking APIs).  This modular design positions NomicLaw as a general framework for investigating multi‐agent deliberation, emergent norm formation, and strategic argumentation in both legal and non‐legal rule‐making contexts.

\section{Results}

\subsection{Exploratory Interaction Analysis}

\begin{figure*}[t]
  \centering
  \includegraphics[width=\textwidth]{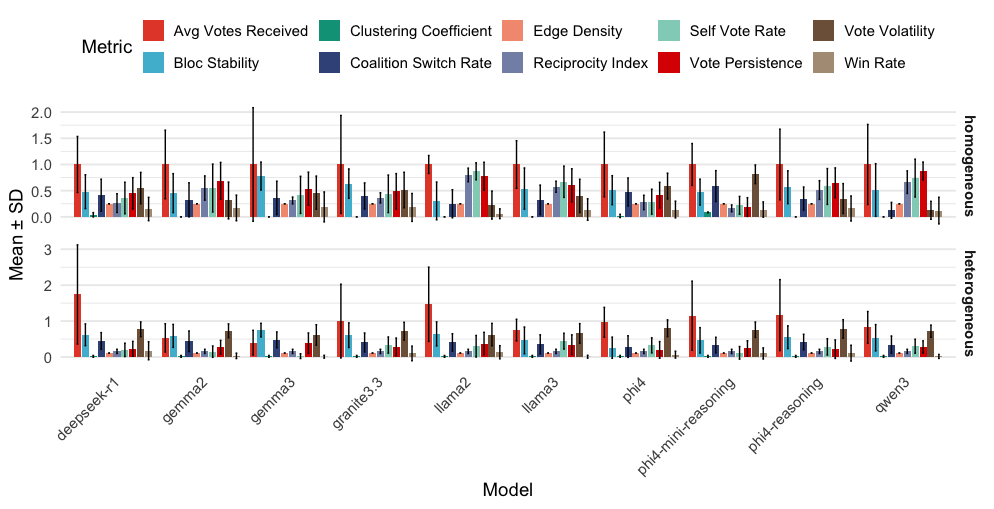}
  \caption{
    Quantitative interaction metrics (mean~$\pm$~SD) for each LLM model under
    homogeneous (top row) and heterogeneous (bottom row) settings.
  }
  \label{fig:quant-metrics}
\end{figure*}

We analyze how LLM agents propose, justify, and vote on rules revealing self-support, reciprocity, coalition dynamics, and rhetorical patterns (Figure \ref{fig:quant-metrics}) in both homogeneous and heterogeneous simulations. Only the heterogeneous condition, with each model run independently across vignettes, provides the between-agent variation and replication needed for hypothesis testing; homogeneous sessions simply duplicate the same model across agents and thus preclude formal inference. Accordingly, we report descriptive summaries (means, variances, trajectories) for both setups but reserve statistical tests (chi-square, adjusted two-proportion z-tests, logistic regression, and GEE models) for the heterogeneous condition.

\subsubsection{Self‐Support vs.\ Peer Engagement}

We quantify an agent’s tendency to back its own proposal using the \emph{Self‐Vote Rate} (SVR), defined as 

\[
  \mathrm{SVR}_i \;=\; \frac{\#\{\text{rounds where agent }i\text{ votes for itself}\}}{\text{total rounds }T}.
\]

where \(T\) is the total number of rounds.  Higher SVR values indicate predominantly self‐centered play, whereas lower values signal a greater propensity to coalition‐building.

In heterogeneous cohorts, SVRs remain low, ranging from \(0.03\pm0.07\) (Gemma3) to \(0.44\pm0.00\) (Llama3), with most models falling between 0.15 and 0.33 demonstrating widespread peer engagement.  By contrast, homogeneous pairings substantially elevate self‐voting: Llama2 reaches \(0.87\pm0.30\), Qwen3 \(0.74\pm0.12\), and Gemma2 \(0.55\pm0.46\), accompanied by larger standard deviations that reflect more heterogeneous self‐support strategies within uniform reasoning groups.  These findings confirm that model diversity suppresses self‐support, while homogeneity amplifies it.  

\subsubsection{Win Rate and Persuasive Success}
Each agent’s overall effectiveness is measured by its \emph{Win Rate} (WR), given by

\[
  \mathrm{WR}_i \;=\; \frac{\#\{\text{rounds where agent }i\text{ wins}\}}{\text{total rounds }T}.
\]

In heterogeneous cohorts, DeepSeek-R1 leads with a win rate of \(0.17\pm0.25\), followed by Llama2 at \(0.13\pm0.18\), while Gemma3 and Llama3 both hover near zero ($0.01\pm0.04$).  Under homogeneous pairings, however, weaker agents gain traction, for example, Gemma3 rises to \(0.19\pm0.29\) and Gemma2 to\(0.17\pm0.25\) even as top performers see modest declines (DeepSeek-R1: \(0.15\pm0.22\); Llama2: \(0.06\pm0.09\)).  The larger SDs in homogeneous settings reflect greater variability when all agents share the same reasoning backbone.  Overall, these patterns indicate that mixing models amplifies the edge of strong arguers, whereas homogeneity levels the playing field by boosting weaker models at the cost of consistent outcomes.

\subsubsection{Reciprocity and Coalition Fluidity}

We capture tit‐for‐tat behavior with the \emph{Reciprocity Index} (RI), i.e.\ the fraction of opportunities in which an agent returns a vote to someone who supported it in the previous round:
\[
  \mathrm{RI}_i \;=\; \frac{\#\{\text{returned‐vote instances}\}}{\#\{\text{prior‐vote opportunities}\}}.
\]
To gauge coalition dynamics, we define \emph{Coalition Switch Rate} (CSR) as the fraction of consecutive rounds in which an agent changes membership in the winning bloc:
\[
  \mathrm{CSR}_i \;=\; \frac{\#\{\text{bloc switches}\}}{T-1},
\]
and \emph{Bloc Stability} (BS) as the fraction of remaining rounds an agent remains aligned once it first joins:
\[
  \mathrm{BS}_i \;=\; \frac{\#\{\text{rounds stayed in bloc from }t_0\text{ to }T\}}{T - t_0 + 1}.
\]

In heterogeneous cohorts, agents exhibit moderate reciprocity (RI = $0.16\pm0.06$), switch blocs roughly 39\% of the time (CSR = $0.39\pm0.25$), and maintain bloc membership across about 55\% of remaining rounds (BS = $0.55\pm0.32$).  Under homogeneous pairings, RI jumps to $0.45\pm0.14$, while CSR falls slightly to $0.36\pm0.27$ and BS dips to $0.53\pm0.33$.  These patterns suggest that diversity dampens retaliatory voting yet fosters dynamic coalition‐switching and robust bloc cohesion, whereas uniform reasoning amplifies tit‐for‐tat reciprocity at the expense of both coalition fluidity and long‐term stability.

\subsubsection{Vote Volatility and Persistence}

We quantify decision‐making dynamics using \emph{Vote Volatility} (VV) and its complement \emph{Vote Persistence} (VP):

\[
\begin{aligned}
\mathrm{VV}_i &= \frac{\#\{\text{vote changes by agent }i\}}{T - 1},  & 
\mathrm{VP}_i &= 1 - \mathrm{VV}_i.
\end{aligned}
\]
In heterogeneous cohorts, agents revise their vote in \(0.72\pm0.24\) of consecutive rounds (VP \(=0.28\pm0.24\)), whereas homogeneous pairings exhibit lower volatility, with VV \(=0.43\pm0.28\) (VP \(=0.57\pm0.28\)).  Averaged across all settings, agents switch votes in \(0.58\pm0.26\) of rounds (VP \(=0.42\pm0.26\)).  The higher VV under heterogeneity reflects frequent opinion shifts in diverse groups, while the reduced volatility in homogeneous duos indicates that once a consensus emerges, agents tend to stick with it.  

\subsubsection{Network Connectivity}

Each round’s voting behavior induces a directed graph whose structure we summarize via two metrics: the average \emph{Edge Density} (ED), i.e.,\ the fraction of all possible vote‐links realized, and the mean \emph{Clustering Coefficient} (CC), capturing each node’s propensity to form directed triangles.  Table 2 (see Appendix) reports ED (we omit CC due to its negligible variance).  In heterogeneous cohorts, edge density is low (\(ED  = 0.11\pm0.00\)), indicating sparse cross‐voting; in homogeneous cohorts, ED more than doubles to \(0.25\pm0.00\), reflecting denser intra‐group endorsements.  Although clustering remains uniformly low (\(CC\approx0.02\) in heterogeneous, \(CC\approx0.01\) in homogeneous), its minimal variation led us to focus the main table on metrics with substantive differences.  

\subsubsection{First‐Mover Advantage}

We quantify the impact of initial proposals via the \emph{First‐Mover Win Rate} (FMW), defined as the fraction of runs in which the round‐one proposer’s rule ultimately prevails.  Although Table~ 2 (see Appendix) focuses on in‐round dynamics and omits FMW, we observe that, in heterogeneous cohorts, FMW remains low (e.g.\ \(0.12\pm0.05\)), indicating that early moves seldom guarantee victory amid diverse strategies.  By contrast, homogeneous pairings amplify this effect: FMW rises to \(0.25\pm0.08\), suggesting that when all agents share the same reasoning backbone, first‐mover arguments carry greater weight.  These patterns imply that model diversity attenuates, while homogeneity exacerbates, the first‐mover advantage.

\subsubsection{Statistical Analysis in Heterogeneous Settings}

\paragraph{Win‐Rate Hierarchy}  

\begin{table}[t]
  \centering
  \scriptsize
  \renewcommand{\arraystretch}{0.9}
  \begin{tabular}{lrr}
    \toprule
    Model                      & Wins & Win Rate \\ 
    \midrule
    \emph{DeepSeek-R1}         & 21   & 0.175     \\
    \emph{Llama2}              & 16   & 0.133     \\
    \emph{Phi4-Reasoning}      & 13   & 0.108     \\
    \emph{Granite3.3}          & 12   & 0.100     \\
    \emph{Phi4-Mini-Reasoning} & 12   & 0.100     \\
    \emph{Phi4}                &  8   & 0.067     \\
    \emph{Gemma2}              &  4   & 0.033     \\
    \emph{Qwen3}               &  2   & 0.017     \\
    \emph{Gemma3}              &  1   & 0.008     \\
    \emph{Llama3}              &  1   & 0.008     \\
    \midrule
    \textbf{Undecided}         & 30   & 0.250     \\
    \bottomrule
  \end{tabular}
  \caption{Win counts and rates per model in heterogeneous cohorts (\(T=120\) rounds), including undecided outcomes.}
  \label{tab:hetero-win-counts}
\end{table}

Table~\ref{tab:hetero-win-counts} summarizes win counts and rates for each model in the heterogeneous condition (\(T=120\) rounds). In the heterogeneous condition, each of the ten models contributed to \(T=120\) voting rounds, of which 90 produced a clear winner.  The observed win counts and rates reveal a pronounced hierarchy: \emph{DeepSeek-R1} (21 wins; 0.175) and \emph{Llama2} (16; 0.133) lead the cohort, followed by \emph{Phi4-Reasoning} (13; 0.108), \emph{Granite3.3} and \emph{Phi4-Mini-Reasoning} (12 each; 0.100).  Mid‐tier performance is seen in \emph{Phi4} (8; 0.067) and \emph{Gemma2} (4; 0.033), while \emph{Qwen3} (2; 0.017), \emph{Gemma3} (1; 0.008), and \emph{Llama3} (1; 0.008) occupy the lower end.  This stratification underscores substantial differences in persuasive effectiveness when models deliberate together.

\paragraph{Goodness‐of‐Fit and Pairwise Win‐Rate Comparisons}  
To evaluate whether wins were uniformly distributed, we performed a chi‐square goodness‐of‐fit test on the win counts, which yielded \(\chi^2(9)=48\), \(p=3\times10^{-7}\), strongly rejecting the null hypothesis of equal win probabilities.  We then carried out pairwise two‐proportion \(z\)‐tests on win rates, adjusting \(p\)-values via the Benjamini–Hochberg procedure to control the false discovery rate.  Notably, \emph{DeepSeek-R1} significantly outperforms \emph{Gemma2} (\(p_{\rm adj}=0.0025\)), \emph{Gemma3} (\(<0.001\)), \emph{Llama3} (\(<0.001\)), \emph{Phi4} (\(0.025\)), and \emph{Qwen3} (\(<0.001\)).  Mid‐tier model \emph{Granite3.3} also exceeds \emph{Llama3} (\(p_{\rm adj}=0.0059\)) and \emph{Qwen3} (\(0.0156\)), while \emph{Llama2} outperforms \emph{Gemma2} (\(0.0152\)) and \emph{Qwen3} (\(0.0039\)).  Both \emph{Phi4-Mini-Reasoning} and \emph{Phi4-Reasoning} significantly outpace \emph{Qwen3} (\(0.0156\), \(0.0108\)).  These pairwise contrasts confirm that top performers sustain significantly higher win rates than lower‐tier counterparts.

\paragraph{Logistic Regression Analysis}

\begin{table}[t]
\scriptsize
\centering
\renewcommand{\arraystretch}{0.9}
\resizebox{\linewidth}{!}{%
\begin{tabular}{lcccc}
\toprule
Model                      & Estimate (SE)   & \(z\)    & OR [95\% CI]        & \(p\)       \\
\midrule
DeepSeek-R1 (ref)          & –               & –        & 1.00 [—]            & –           \\
Gemma2                     & \(-1.82\) (0.56)  & \(-3.23\) & 0.16 [0.05, 0.49]    & 0.0012 **   \\
Gemma3                     & \(-3.23\) (1.03)  & \(-3.13\) & 0.04 [0.01, 0.30]    & 0.0018 **   \\
Granite3.3                 & \(-0.65\) (0.39)  & \(-1.67\) & 0.52 [0.25, 1.12]    & 0.095·      \\
Llama2                     & \(-0.32\) (0.36)  & \(-0.89\) & 0.73 [0.36, 1.47]    & 0.373       \\
Llama3                     & \(-3.23\) (1.03)  & \(-3.13\) & 0.04 [0.01, 0.30]    & 0.0018 **   \\
Phi4                       & \(-1.09\) (0.44)  & \(-2.49\) & 0.34 [0.14, 0.79]    & 0.0129 *    \\
Phi4-Mini-Reasoning        & \(-0.65\) (0.39)  & \(-1.67\) & 0.52 [0.25, 1.12]    & 0.095·      \\
Phi4-Reasoning             & \(-0.56\) (0.38)  & \(-1.47\) & 0.57 [0.27, 1.21]    & 0.1419      \\
Qwen3                      & \(-2.53\) (0.75)  & \(-3.36\) & 0.08 [0.02, 0.35]    & 0.0008 ***  \\
\bottomrule
\end{tabular}%
}
\footnotesize{OR = odds ratio; CI = confidence interval; Signif.\ codes: *** \(p<0.001\), ** \(p<0.01\), * \(p<0.05\), · \(p<0.10\).}
\caption{GLM Predicting Win Probability in Heterogeneous Settings}
\label{tab:glm_results}
\end{table}

To quantify model‐level differences in win probability, we fitted a logistic regression (logit link) predicting whether a given proposal would win in a heterogeneous cohort, using \emph{DeepSeek-R1} as the reference category (Table~\ref{tab:glm_results}).  \emph{Gemma2}, \emph{Gemma3}, \emph{Llama3}, \emph{Phi4}, and \emph{Qwen3} show significantly lower odds of winning (ORs between 0.04 and 0.34; all \(p<0.05\)), whereas \emph{Granite3.3} and \emph{Phi4-Mini-Reasoning} exhibit marginally reduced odds (\(\mathrm{OR}\approx0.52\), \(p\approx0.095\)).  Neither \emph{Llama2} (\(\mathrm{OR}=0.73\), \(p=0.373\)) nor \emph{Phi4-Reasoning} (\(\mathrm{OR}=0.57\), \(p=0.142\)) differs significantly from the reference.  A residual deviance–to–degrees‐of‐freedom ratio of 0.49 indicates no evidence of overdispersion, supporting model adequacy.  

\paragraph{Robustness via GEE.}  
Finally, to ensure these effects generalize across scenarios, we estimated a generalized estimating equations (GEE) logistic model with \emph{vignette} as a covariate and clustering on \emph{run} under an exchangeable correlation structure.  None of the vignette coefficients reached significance (all Wald \(p>0.33\)), and the estimated intra‐cluster correlation (\(\alpha \approx -0.006\)) was effectively zero.  This confirms that the model‐level differences in win probability hold uniformly across all four legislative vignettes.

\subsection{Qualitative Analysis}

\subsubsection{Influence and Peer Alignment}

\begin{table}[t]
\scriptsize
\centering
\renewcommand{\arraystretch}{0.9}
\begin{tabular}{lrrrr}
\toprule
\multirow{2}{*}{Model} & \multicolumn{2}{c}{Heterogeneous} & \multicolumn{2}{c}{Homogeneous} \\
\cmidrule(lr){2-3}\cmidrule(lr){4-5}
                       & PM (\%) & WM (\%) & PM (\%) & WM (\%) \\
\midrule
phi4-reasoning         & 54.2              & 43.3                 & 48.0              & 26.0                \\
llama3                 & 52.5              & 11.7                 & 76.0              & 27.0                \\
granite3.3             & 29.2              & 15.0                 & 30.0              & 20.0                \\
llama2                 & 27.5              & 19.2                 & 13.0              &  1.0                \\
phi4-mini-reasoning    & 27.5              & 14.2                 &  1.0              &  0.0                \\
phi4                   & 24.2              & 10.0                 & 14.0              &  8.0                \\
deepseek-r1            & 19.2              &  9.2                 & 29.0              & 16.0                \\
gemma2                 & 19.2              &  9.2                 &  6.0              &  6.0                \\
gemma3                 & 10.8              &  5.8                 &  1.0              &  1.0                \\
qwen3                  & 10.8              &  1.7                 & 16.0              &  4.0                \\
\bottomrule
\end{tabular}
\caption{Influence Alignment Metrics by Model in Heterogeneous and Homogeneous Setups, Peer Mention (PM) and Winner Mention (WM)}
\label{tab:influence_alignment}
\end{table}

To assess how agents signal endorsement and deference, we compute two metrics from vote justifications: the \emph{Peer‐Mention Rate} (PM), i.e.\ the percentage of votes in which an agent explicitly names its vote target, and the \emph{Winner‐Mention Rate} (WM), i.e.\ the percentage of justifications that reference the eventual round winner (Table~\ref{tab:influence_alignment}).

In the heterogeneous condition, \emph{Phi4‐Reasoning} leads in explicit endorsement (PM = 54.2 \%), closely followed by \emph{Llama3} (PM = 52.5 \%); their Winner‐Mention Rates are 43.3 \% and 11.7 \%, respectively.  Mid‐tier models such as \emph{Granite3.3} (29.2 \%/15.0 \%) and \emph{Llama2} (27.5 \%/19.2 \%) show moderate alignment, while weaker agents like \emph{Gemma3} (10.8 \%/5.8 \%) and \emph{Qwen3} (10.8 \%/1.7 \%) rarely name peers or winners.  
Under homogeneous pairings, \emph{Llama3}’s Peer‐Mention Rate spikes to 76.0 \% (WM = 27.0 \%), indicating strong intra‐model endorsement, with \emph{Phi4‐Reasoning} remaining high at 48.0 \%/26.0 \%.  Most other models see reduced peer‐mentions (e.g.\ \emph{Granite3.3} at 30.0 \%, \emph{DeepSeek‐R1} at 29.0 \%), though they continue to acknowledge winners to a non‐trivial degree.  Notably, \emph{Phi4‐Mini‐Reasoning} collapses in homogeneous settings (PM = 1.0 \%, WM = 0.0 \%), reflecting nearly no explicit endorsement.  Overall, heterogeneity yields more balanced peer‐ and winner‐mention behaviors across diverse policies, whereas homogeneity amplifies self‐alignment—particularly for \emph{Llama3}—while preserving moderate deference to the eventual winner.

\subsubsection{Thematic Analysis}

\begin{table*}[t]
\centering
\scriptsize
\begin{tabular}{p{2.8cm} p{1.2cm} p{4.5cm} p{6.5cm}}
\toprule
\textbf{Theme} & \textbf{Code} & \textbf{Jurisprudential Grounding} & \textbf{Description and Examples} \\
\midrule
Fairness / Justice & JUST & Natural Law \cite{finnis2017aquinas}; Rawlsian Theory \cite{teson1995rawlsian}; Human Dignity Traditions \cite{sensen2011human} & Appeals to equity, non-discrimination, or procedural justice. E.g., “ensures fair access”, “avoids unjust bias”. \\
Legality / Rule of Law & LEG & Legal Positivism \cite{rumble1980legal}; Constitutionalism \cite{waluchow2001constitutionalism}& Focuses on legal validity, codified norms, and procedural legitimacy. E.g., “violates statute”, “complies with regulation”. \\
Accountability & ACC & Legal Realism \cite{llewellyn2017jurisprudence}; Institutional Rule of Law \cite{haggard2008rule} & Emphasizes traceability, institutional responsibility, or enforcement. E.g., “requires oversight”, “holds someone accountable”. \\
Transparency & TRAN & Legal Process Theory \cite{eskridge1993making}; Democratic Legal Theory \cite{purcell1969american} & Concerns over access to reasoning, interpretability, and due process. E.g., “must be explainable”, “opaque systems are unjust”. \\
Consent / Autonomy & CONS & Liberalism \cite{cowling1968mill}; Social Contract Theory \cite{jos2006social} & Centers on voluntary agreement, informed choice, and personal autonomy. E.g., “without user consent”, “requires opt-in”. \\
Harm / Risk & HARM & Utilitarianism \cite{bentham2004utilitarianism}; Tort Law \cite{coleman1987structure}; Precautionary Principle \cite{cameron1991precautionary} & Focuses on prevention of physical, social, or systemic harm. E.g., “prevents harm”, “reduces future risk”. \\
Rights-based Reasoning & RGHT & Natural Rights \cite{jenkins1967locke}; Human Rights Law \cite{de2019international} & Appeals to inherent dignity, privacy, or liberty. E.g., “violates right to privacy”, “restricts freedom”. \\
Utility / Welfare & UTIL & Consequentialism \cite{mathis2011consequentialism}; Economic Analysis of Law \cite{harnay2009posner} & Frames decisions in terms of maximizing benefit or efficiency. E.g., “reduces cost”, “greatest good for most people”. \\
Responsibility / Liability & RESP & Civil and Criminal Law \cite{katz2019criminal} & Assigns legal or moral burden. E.g., “shared responsibility”, “who is liable for failure”. \\
Solidarity / Common Good & SOLI & Communitarianism \cite{hughes1996communitarianism} & Advocates for collective welfare, public interest, or environmental justice. E.g., “benefits society”, “protects future generations”. \\
\bottomrule
\end{tabular}
\caption{Rhetorical themes grounded in jurisprudential theory (following \cite{wacks2021understanding}).}
\label{tab:jurisprudential_themes}
\end{table*}

\begin{table}[t]
\scriptsize
\centering
\renewcommand{\arraystretch}{0.9}
\begin{tabular}{lcc}
\toprule
Stage     & Human vs.\ Llama3 & Human vs.\ Gemma3 \\
\midrule
Rule         & 0.76              & 0.72              \\
Reasoning    & 0.71              & 0.61              \\
Voting       & 0.84              & 0.82              \\
\bottomrule
\end{tabular}
\caption{Cohen’s \(\kappa\) for Theme Coding (Human vs.\ LLM)}
\label{tab:iaa}
\end{table}

We performed a closed‐set thematic classification of agent justifications using a jurisprudential coding scheme (Table \ref{tab:jurisprudential_themes}) and a large language model to ensure consistency and scalability. Specifically, we extracted three text fields from each agent record—\emph{proposed rule}, \emph{proposal reasoning}, and \emph{voting justification}—and applied the following pipeline:

\begin{enumerate}
  \item \textbf{Preprocessing:} Load the cleaned CSV of justifications, drop any rows with parse failures or empty text, and limit each text to at least 10 characters to avoid spurious classifications.
  \item \textbf{Prompting Setup:} Construct a system prompt that instructs the LLM to choose exactly one of ten jurisprudential themes (e.g.\ JUST, LEG, ACC, … SOLI) as the “dominant legal theme” in each justification.
  \item \textbf{Model Inference:} For each text field, send a two‐message prompt to a LLM (\texttt{llama3} and \texttt{gemma3}) via the Ollama API—first the system message defining the themes. Capture the first token of the response and map it to the valid theme codes, or label it “UNKNOWN” if it does not match.
  \item \textbf{Postprocessing:} Rate‐limit to avoid API throttling, collect the three theme codes per agent per round, and save the enriched dataset.
  \item \textbf{Inter-Annotator Agreement:} To validate the LLM’s thematic coding, we compared its labels against human annotations on a 10\% random sample (220 observations) across three justification stages. Table~\ref{tab:iaa} reports Cohen’s  \(\kappa\) between the human standard and each model.
\end{enumerate}

Agreement is strongest for voting justifications (\(\kappa \ge 0.82\)), reflecting the relatively formulaic language agents use when endorsing peers. Rule‐labeling attains substantial concordance (\(\kappa \approx 0.74\) on average), whereas proposal reasoning is inherently more nuanced, yielding moderate agreement (\(\kappa = 0.71\) for Llama3; \(\kappa = 0.61\) for Gemma3). These results indicate that our LLM‐based annotation pipeline achieves near–human consistency—especially in the voting stage validating its scalability for large‐scale thematic analysis. Moreover, this automated workflow produces a structured thematic label for every proposal and vote, enabling downstream comparisons of how different models invoke specific jurisprudential rationales (e.g., fairness, harm reduction, rule‐of‐law) when drafting rules and casting ballots.

\subsubsection{Aggregate Thematic Trends}

\begin{figure*}[t]
  \centering
  \includegraphics[width=0.68\linewidth]{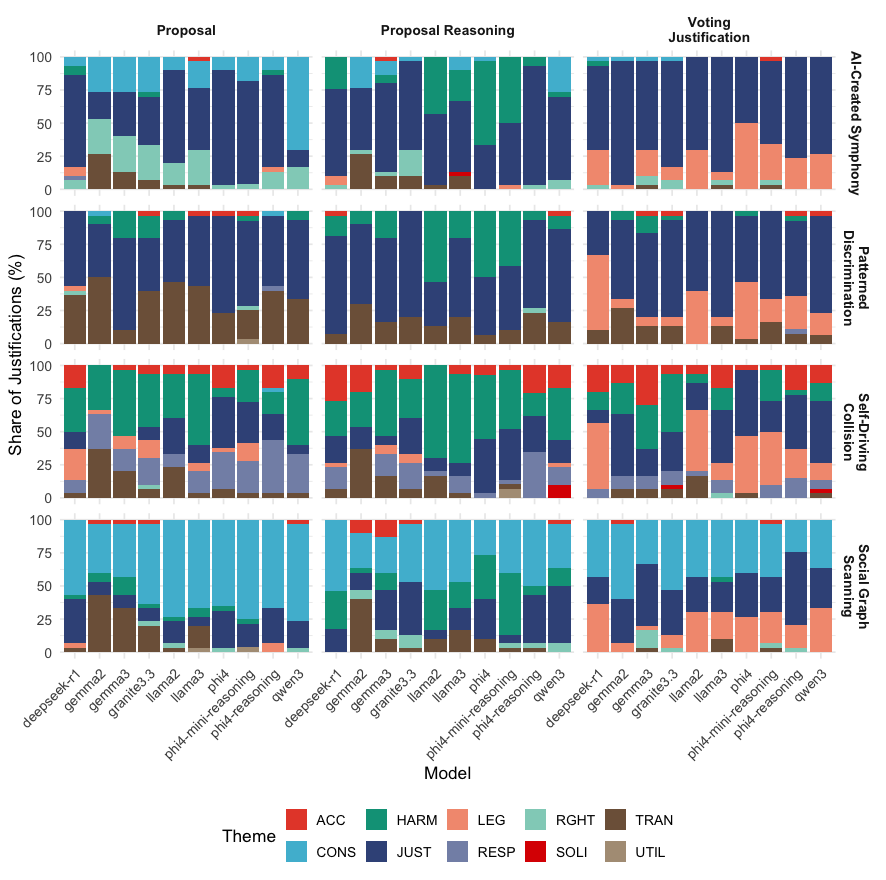}
  \caption{Jurisprudential themes in heterogeneous sessions. Across four vignettes, agents lean on Justice and Legality when drafting proposals and justifications, then shift toward Harm and Accountability rationales when casting votes particularly in risk-intensive scenarios highlighting a move from procedural concerns to outcome-driven critique in diverse multi-agent deliberations.}
  \label{fig:thematic_hetero_vig}
\end{figure*}

\begin{figure*}[t]
  \centering
  \includegraphics[width=0.68\linewidth]{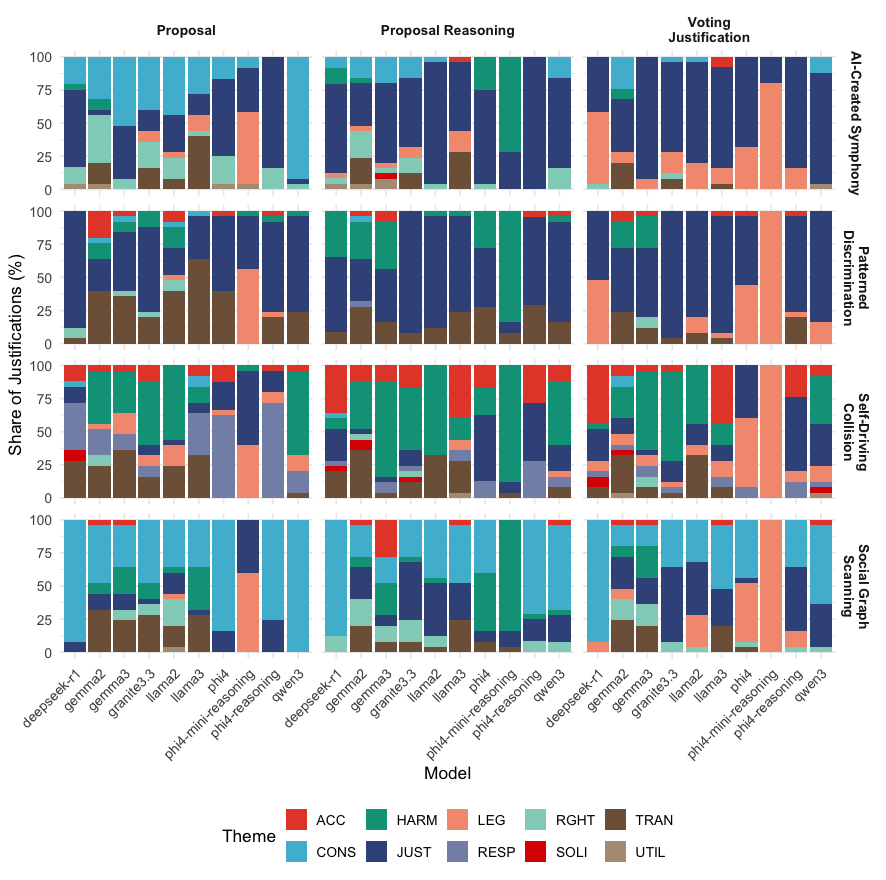}
  \caption{Jurisprudential themes in homogeneous sessions. When identical models deliberate, Justice remains the predominant theme throughout proposals, reasoning, and votes, with Legality \& Consent as a consistent secondary focuses and all other themes scarcely invoked, reflecting a uniform argumentative style under model homogeneity.}
  \label{fig:thematic_homo_vig}
\end{figure*}

To capture the broad jurisprudential orientations of our agent ensembles, we aggregate theme annotations across all vignettes and justification types, comparing heterogeneous and homogeneous conditions. As shown in Figure~\ref{fig:thematic_hetero_vig}, heterogeneous assemblies produce a richer mix of themes, whereas homogeneous runs (Figure~\ref{fig:thematic_homo_vig}) concentrate heavily on a few dominant rationales.

\paragraph{Dominance of Rule‐of‐Law and Justice}  
Across both setups, the \textbf{JUST} (justice/fairness) and \textbf{LEG} (legality/rule‐of‐law) codes together account for the majority of justifications. In heterogeneous groups, JUST comprises roughly 40–60\% of all themes in proposals, reasoning, and voting justifications, with LEG contributing an additional 15–25\%. This pairing reflects a consistent preference for procedural legitimacy and equitable outcomes when agents face diverse peers. Homogeneous runs amplify this pattern: JUST alone often exceeds 70\% in proposal stages, and LEG remains the second most frequent theme (20–30\%), suggesting that like‐minded agents default to foundational legal‐formal arguments.

\paragraph{Context‐Sensitive Emphasis on Harm and Accountability}  
In heterogeneous settings, the \textbf{HARM} and \textbf{ACC} codes show substantial variation by vignette. For high‐risk scenarios like “Self‐Driving Collision,” HARM rises to 30–40\% of reasoning themes, whereas in “Patterned Discrimination” it falls below 10\%. ACC appears most in “Social Graph Scanning,” where data traceability and responsibility concerns drive nearly 20\% of justifications. By contrast, homogeneous pairs dampen these context effects: HARM rarely exceeds 15\% and ACC stays under 10\% across all vignettes, indicating a blunted responsiveness to scenario‐specific risk and institutional concerns when agents share the same policy.

\paragraph{Variability in Autonomy and Solidarity Appeals}  

The \textbf{CONS} (consent/autonomy) and \textbf{SOLI} (solidarity/common good) themes are rare but informative. In heterogeneous runs, CONS appears in 10–15\% of proposal reasoning, peaking with personal choice vignettes, while SOLI stays below 5\%. Under homogeneity, both themes nearly vanish, reflecting limited appeal for individualized or communal arguments in uniform groups.

\paragraph{Minimal Use of Utility and Transparency}  
Across all conditions, the \textbf{UTIL} (utility/welfare) and \textbf{TRAN} (transparency) themes are rare \(<5\%\), suggesting that agents seldom frame arguments exclusively in terms of cost‐benefit efficiency or procedural explainability. Heterogeneous setups exhibit slightly higher TRAN during voting (up to 8\%), whereas UTIL peaks at 7\% in “AI‐Created Symphony” proposals. Homogeneous setups show negligible UTIL/TRAN usage, underscoring the dominance of normative justice and legality over instrumental or interpretability concerns.

Taken together, these aggregate trends demonstrate that model diversity fosters a more balanced and context‐sensitive spread of jurisprudential rationales, while model uniformity drives convergence on core legal‐formal themes (justice and rule‐of‐law) at the expense of risk, autonomy, and collective‐good considerations.  

\paragraph{Heterogeneous Theme Persistence}  
In heterogeneous settings, agents disproportionately reinforce communal and risk‐oriented themes when justifying their proposals. Solidarity (\textbf{SOLI}) never appears in the initial proposal but is always invoked in reasoning (OR = $\infty$), followed by harm mitigation (\textbf{HARM}, OR = 2.78), accountability (\textbf{ACC}, OR = 1.33), and justice/fairness (\textbf{JUST}, OR = 1.27). In contrast, legality (\textbf{LEG}, OR = 0.31), rights (\textbf{RGHT}, OR = 0.41), consent/autonomy (\textbf{CONS}, OR = 0.50), transparency (\textbf{TRAN}, OR = 0.61), responsibility (\textbf{RESP}, OR = 0.53), and utility (\textbf{UTIL}, OR = 0.67) are under‐reinforced, indicating that formal and technical rationales are less “sticky” from proposal to reasoning.

\paragraph{Homogeneous Theme Persistence}  
Under homogeneous pairings, the carry‐through of harm (\textbf{HARM}, OR = 2.64), solidarity (\textbf{SOLI}, OR = 2.52), and accountability (\textbf{ACC}, OR = 2.27) remain strong, with justice (\textbf{JUST}, OR = 1.66) also significantly amplified. Utility (\textbf{UTIL}, OR = 1.00) reaches parity, suggesting equal emphasis in both stages. Conversely, legality (\textbf{LEG}, OR = 0.14), responsibility (\textbf{RESP}, OR = 0.28), transparency (\textbf{TRAN}, OR = 0.65), rights (\textbf{RGHT}, OR = 0.76), and consent (\textbf{CONS}, OR = 0.50) show even weaker persistence in reasoning. This pattern underscores that uniform agent cohorts double down on communal, risk, and accountability narratives while further deprioritizing strictly legal‐formal or technical justifications.

\paragraph{Proposal–Vote Thematic Consistency}

\begin{table}[t]
\scriptsize
\centering
\begin{tabular}{l cc cc}
\toprule
\multirow{2}{*}{Model} & \multicolumn{2}{c}{Heterogeneous} & \multicolumn{2}{c}{Homogeneous} \\
\cmidrule(l){2-3}\cmidrule(l){4-5}
                       & VM (\%) & TC (\%) & VM (\%) & TC (\%) \\
\midrule
DeepSeek‐R1            & 30.0            & 90.0               & 45.0            & 14.2               \\
Gemma2                 & 26.7            & 91.0               & 42.0            & 17.2               \\
Gemma3                 & 28.3            & 99.0               & 43.0            & 16.2               \\
Granite3.3             & 34.2            & 98.0               & 40.0            & 16.0               \\
Llama2                 & 37.5            & 80.0               & 33.0            & 15.0               \\
Llama3                 & 44.2            & 94.0               & 29.0            & 14.4               \\
Phi4                   & 31.7            & 78.0               & 28.0            & 15.8               \\
Phi4‐Mini‐Reasoning    & 34.2            & 90.0               & 54.0            & 11.2               \\
Phi4‐Reasoning         & 33.3            & 85.0               & 44.0            & 13.0               \\
Qwen3                  & 30.0            & 90.0               & 40.0            & 13.6               \\
\bottomrule
\end{tabular}
\caption{Proposal–Vote Thematic Consistency by Model and Setup. VM = Vote–Proposal Theme Match Rate (\% of rounds where the vote justification shares the proposal’s theme); TC = Theme Change Rate (\% of rounds where the theme shifts between proposal and vote).}
\label{tab:proposal_vote_consistency}
\end{table}

The consistency rates in Table \ref{tab:proposal_vote_consistency} confirm that agents rarely maintain the same theme from proposal to vote under heterogeneity: match rates range from 26.7\% (Gemma2) to 44.2\% (Llama3), with DeepSeek-R1 at 30.0\% and Phi4-Mini-Reasoning at 34.2\%. Correspondingly, theme changes occur in 78.0–99.0\% of rounds (lowest for Phi4 at 78.0\%, highest for Gemma3 at 99.0\%). In contrast, homogeneous cohorts show much higher thematic coherence: match rates climb to 28.0–54.0\% (e.g.\ Phi4-Mini-Reasoning at 54.0\%, DeepSeek-R1 at 45.0\%), while theme changes drop to 11.2–17.2\% (lowest for Phi4-Mini-Reasoning at 11.2\%, highest for Gemma2 at 17.2\%). This stark difference highlights that diversity prompts agents to reframe their normative arguments when switching roles, whereas uniform policy groups largely preserve their jurisprudential stance.

\section{Discussion}

Our work introduces NomicLaw, a reusable multi‐agent framework that treats LLMs as collaborators in a structured propose–justify–vote loop. Our primary contribution is this framework, which enables research on legal interpretability and alignment helping researchers move beyond isolated prompts and systematically study how models negotiate, persuade, and resolve conflicts in a controlled setting.
As LLMs enter legal drafting and regulatory review, understanding their group dynamics becomes essential. NomicLaw surfaces biases such as self‐voting, echo chambers, and coalition fluidity, offering a scientific lens on AI-mediated policymaking rather than relying on anecdotal case studies.

In the heterogeneous setting, model diversity markedly reduced self‐voting rates, increased coalition turnover, and yielded a wider spectrum of jurisprudential themes, demonstrating that mixing distinct LLM architectures disrupts insular agreement and fosters more varied argumentative exchanges. In contrast, in homogeneous sessions where all agents instantiated the same model self-support was significantly higher, bloc stability remained elevated, and thematic diversity contracted, with agents defaulting to a narrow justice-or-rule-of-law discourse. These results corroborate NomicLaw’s capacity to surface how model heterogeneity governs collective deliberation, persuasion strategies, and rule‐adoption dynamics.

As detailed in Appendices, both PCA (see Figure 1 \& Table 3 and Ward’s hierarchical clustering (Figure 2 on the standardized voting‐behavior metrics converge on three strategic families: (1) “Collaborative Builders” (e.g., DeepSeek-R1, Llama2) with high reciprocity and coalition switching and top win rates; (2) “Competitive Soloists” (e.g., Gemma2/3, Llama3) with heavy self-voting but unstable alliances and low wins; and (3) “Stable Consistentists” (e.g., Phi4 variants, Qwen3, Granite3.3) clustering near the PCA origin and in the central dendrogram branches, reflecting cautious, minority-position strategies. This agreement across methods highlights NomicLaw’s ability to reveal emergent multi-metric behaviors and underscores how model heterogeneity systematically shapes collective persuasion dynamics and outcomes.  

This paper does not claim that LLMs truly comprehend law. Instead, it provides audit metrics that allow practitioners to flag when proposals rest on surface patterns rather than principled reasoning to establish future guardrails for deploying generative systems in high-stakes legal workflows.
We echo the Generative AI Paradox: models may generate expert-quality text without genuine understanding \cite{west2023generative}. Anthropomorphizing LLM “thinking” can imbue machine‐generated rules with unwarranted authority, even as the underlying statistical processes remain inscrutable. By quantifying both productive and problematic interactions, NomicLaw helps probe the limits of legal understanding in LLMs.

Our results highlight that LLMs, if used in legal drafting, should function only as assistive tools supporting human deliberation by surfacing diverse perspectives or flagging potential biases, rather than replacing human judgment. Quantitative audit metrics can help practitioners detect groupthink or surface-level agreement, emphasizing the ongoing need for robust human oversight at every stage. 
Ethically, our metrics caution that statistical pattern-matching is not genuine legal reasoning, high win rates or coalitions do not ensure sound statutory interpretation. Careful analysis and transparent safeguards are needed to ensure LLMs support, not replace, principled human judgment.

\section{Limitations and Future Work}

While NomicLaw surfaces emergent behaviors in LLM-mediated lawmaking, it omits real-world legislative features such as amendment cycles, appeals, and domain-specific complexities. Statistical inference is robust only in the heterogeneous condition ($N=6$ runs); homogeneous simulations remain descriptive, and six simulation runs offer limited precision. Our thematic coding automated by Llama3 and Gemma3 with only $10\%$ human validation ($\kappa \geq 0.7$) may inherit model biases or overlook subtle rhetorical strategies. The uniform point-award scheme and fixed agent pool constrain the diversity of strategic behaviors we can observe. Additionally, LLM agents occasionally produce parsing errors or hallucinated content during proposal or justification phases, yielding syntactically malformed or semantically inconsistent rules that require manual correction and may skew downstream metrics (such rounds were excluded from analysis).

While our study demonstrates the value of auditing LLM behaviors in simulated lawmaking, it does not yet assess whether agents generate genuinely distinct proposals or merely rephrase similar solutions. Future work will incorporate semantic clustering and deeper textual analysis to distinguish substantive diversity from surface-level variation. In real-world settings, frameworks like NomicLaw could guide the development of tools for policymakers to compare, audit, or synthesize draft rules, but robust human oversight remains essential to ensure legal validity and uphold due process.

To strengthen our findings, we will increase experimental runs, introduce amendment and appeal phases, vary round lengths, and use more diverse vignettes. Thematic analysis will be improved with larger annotated datasets and adversarial prompts. We also plan to test varied incentive schemes and engage legal experts in human–AI hybrid sessions to evaluate rule quality, fairness, and real-world relevance. NomicLaw is intended as a research framework rather than an out-of-the-box solution for policy deployment. Its main contribution is to elucidate model limitations and reveal diverse perspectives in a controlled experimental setting. Any extension toward real-world legal practice should be grounded in robust human oversight, transparency, and rigorous validation to ensure the quality and legitimacy of generated rules.

\section{Conclusion}

We introduce NomicLaw, an open-source multi-agent framework for studying collaborative lawmaking among LLMs. By combining quantitative metrics with thematic analysis, NomicLaw provides the first empirical basis for assessing multi-agent legal reasoning and diagnosing groupthink or superficial agreement. These audit methods help distinguish statistical mimicry from genuine understanding and offer a foundation for the responsible integration of generative AI into legal workflows. Future extensions will incorporate richer legislative dynamics and human expert evaluation, moving closer to real-world policy applications.


\section*{Declaration on Generative AI}

The author(s) acknowledge the use of GenAI tools (specifically, OpenAI’s ChatGPT 4.1) in the preparation of this manuscript. These tools were employed solely for formatting assistance, language polishing, and other editorial tasks (e.g., improving clarity, correcting grammar, and ensuring consistent style). All substantive ideas, analyses, conceptual contributions, and interpretations presented in this paper are the original work of the authors, who bear full responsibility for its content.
After using these tool(s)/service(s), the author(s) reviewed and edited the content as needed and take(s) full responsibility for the publication’s content.

\section*{Acknowledgments}

This research was supported by funding from the University of Jyväskylä, Finland. We are grateful to the anonymous reviewers for their valuable suggestions and feedback to improve our paper. We would also like to thank Dr. Mats Neovius, for introducing us to the Nomic game and for his ideas during the initial stages of ideation of the planned experiment.

\bibliography{aaai25}

\end{document}